\documentclass[letterpaper, 10 pt, conference]{ieeeconf}  % Comment this line out if you need a4paper

\IEEEoverridecommandlockouts                              % This command is only needed if 
\usepackage{cuted}
\usepackage{amsmath}
\usepackage{hyperref}
\usepackage{pifont}
\usepackage{marvosym}
\usepackage{amssymb}
\usepackage{subcaption}
\usepackage{booktabs}
\usepackage{multirow}
\usepackage{graphicx}
\usepackage{xcolor}
\usepackage{placeins}
\usepackage[table]{xcolor}

\title{\LARGE \bf
SUM-AgriVLN: Spatial Understanding Memory for\\Agricultural Vision-and-Language Navigation
}

\author{Xiaobei Zhao\textsuperscript{\rm 1,2}\textsuperscript{\Letter}, Xingqi Lyu\textsuperscript{\rm 1}, Xin Chen\textsuperscript{\rm 1,2}\textsuperscript{\Letter}, and Xiang Li\textsuperscript{\rm 1,2}\textsuperscript{\Letter}\thanks{\textsuperscript{\Letter} Corresponding authors.}\normalsize\\ \\
\textsuperscript{\rm 1}China Agricultural University\\
\textsuperscript{\rm 2}China Agricultural University-Sichuan Advanced Agricultural and Industrial Institute\\
\texttt{\{xiaobeizhao2002,lxq99725\}@163.com, \{chxin,cqlixiang\}@cau.edu.cn}}
\begin{document}

\maketitle
\thispagestyle{empty}
\pagestyle{empty}

%%%%%%%%%%%%%%%%%%%%%%%%%%%%%%%%%%%%%%%%%%%%%%%%%%%%%%%%%%%%%%%%%%%%%%%%%%%%%%%%
% enabling robots to navigate to the target positions following the natural language instructions. 
% employs spatial understanding and save spatial memory through 3D reconstruction and representation.
\begin{abstract}
Agricultural robots are emerging as powerful assistants across a wide range of agricultural tasks, nevertheless, they are still heavily relying on manual operations or fixed railways for movement. The A2A benchmark and the AgriVLN method pioneeringly extended Vision-and-Language Navigation (VLN) to the agricultural domain, successfully navigating agricultural robots from starting points to target positions following natural language instructions, while we observed a limitation: In practical agricultural scenarios, users often give repetitive instructions, but AgriVLN treats every instruction as an independent episode, overlooking the potential to use past spatial memories to assist present episodes. To address this limitation, we propose the SUM module, which executes spatial understanding via 3D reconstructions and saves spatial memories via 2D representations from the past, thereby assisting the decision-maker to recall the spatial characteristics of the scenes in the present. We integrate it into the AgriVLN backbone to build the SUM-AgriVLN method. When evaluated on A2A, it effectively improves SR from 0.47 to 0.54 with only slight sacrifice on NE from 2.91 m to 2.93 m, demonstrating the state-of-the-art performance in the agricultural VLN domain. 
Code: \href{https://github.com/AlexTraveling/SUM-AgriVLN}{https://github.com/AlexTraveling/SUM-AgriVLN}.
\end{abstract}
% only
% demonstrating
% Success Rate (SR)
% Navigation Error (NE)
% To address this limitation, we propose the SUM-AgriVLN method, in which the SUM module employs spatial understanding via 3D reconstructions and saves spatial memories via 2D representations.

\begin{figure}[!t]
\centering
  \begin{subfigure}[t]{1.0\linewidth}
    \includegraphics[width=\linewidth]{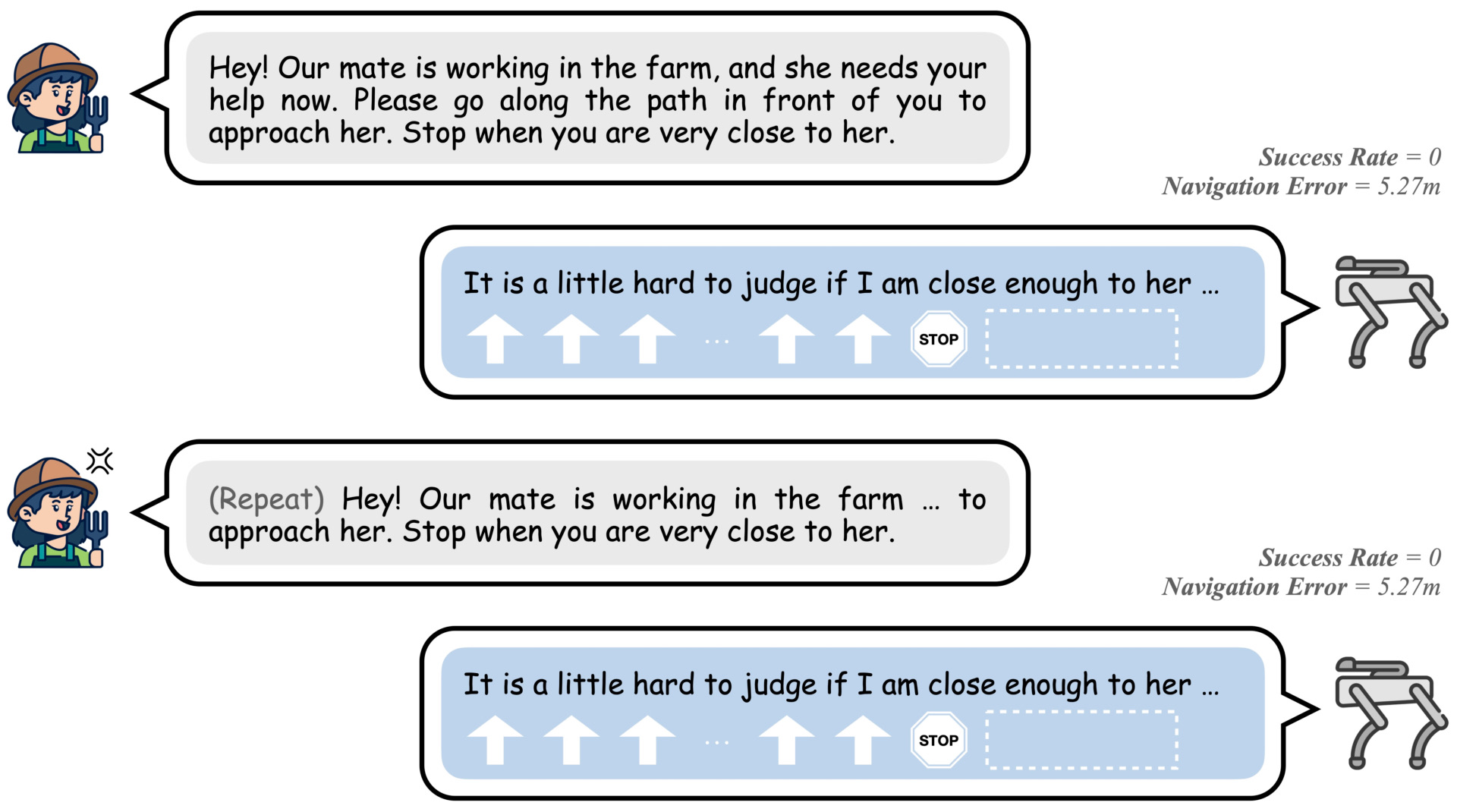}
    \caption{Standard agricultural VLN methods.}
  \end{subfigure}
  \hfill
  \vspace{1.8ex}
  \begin{subfigure}[t]{1.0\linewidth}
    \includegraphics[width=\linewidth]{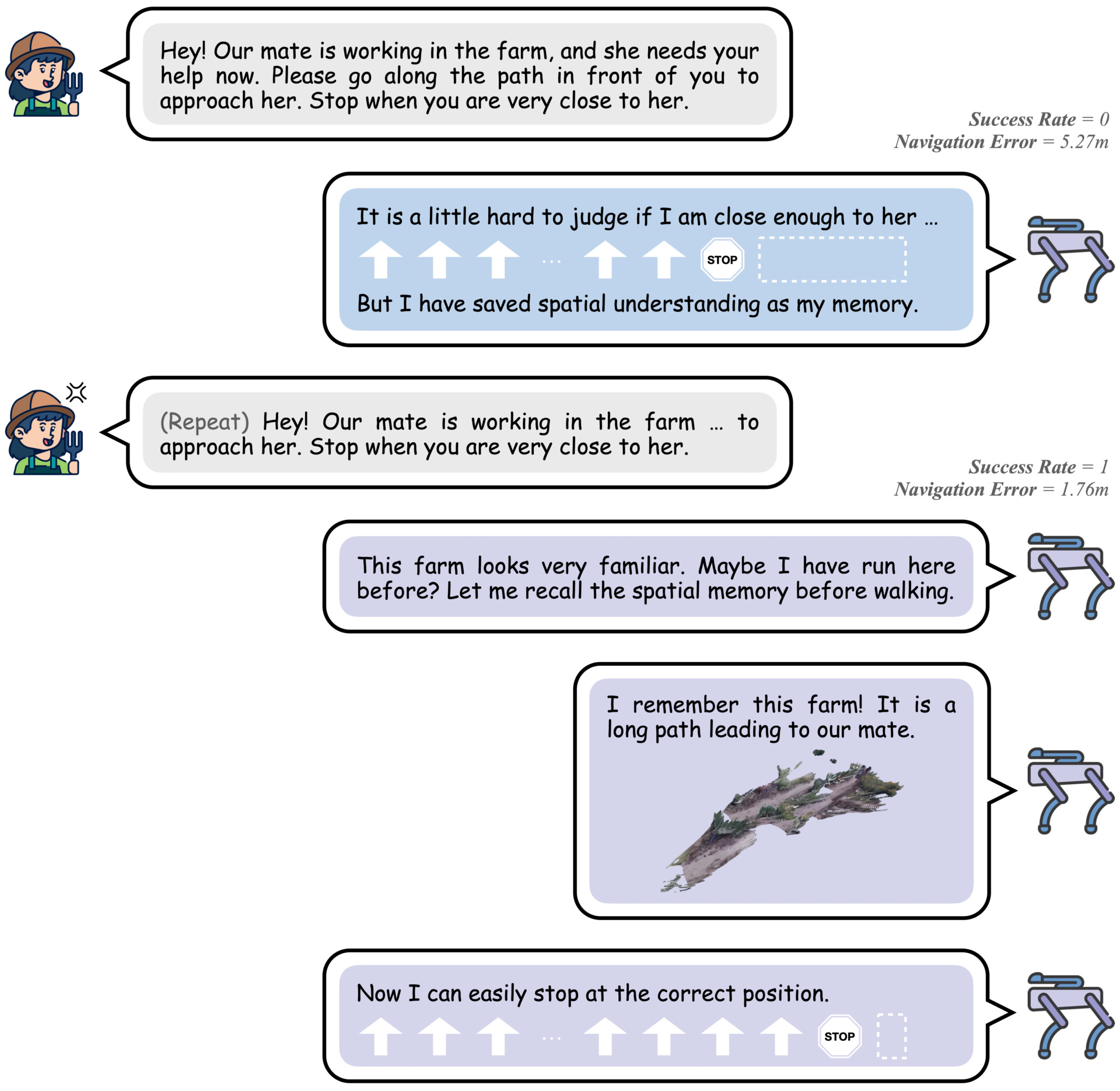}
    \caption{SUM-AgriVLN.}
  \end{subfigure}
\caption{SUM-AgriVLN v.s. standard agricultural VLN methods on a simple example.}
\label{fig:teaser}
\vspace{-0.5cm}
\end{figure}

% Section 1
\section{Introduction}
\par Agricultural robots are emerging as powerful assistants across a wide range of agricultural tasks, such as 
pesticide spraying \cite{JournalFieldRobotics}, 
phenotypic measurement \cite{ICIC:TomatoScanner}, 
and fruit harvesting \cite{ICRA:AHPPEBot}. However, most of them are still heavily relying on manual operations or fixed railways for movement, which results in limited mobility and poor adaptability. 
% untransportable
% --- Submit ---
% \par In contrast, Vision-and-Language Navigation (VLN) enables agents follow the natural language to navigate to the target position \cite{CVPR:R2R,ECCV:VLN-CE}, having demonstrated strong performance across various domains \cite{TMLR}, such as R2R for indoor room \cite{CVPR:R2R}, TOUCHDOWN \cite{CVPR:TOUCHDOWN} for urban street, and AerialVLN \cite{ICCV:AerialVLN} for aerial space. Motivated by prior Vision-Language Model-based studies \cite{AAAI:NavGPT,ACL:MapGPT,ICRA:Discuss}, the AgriVLN \cite{arXiv:AgriVLN} method and the A2A \cite{arXiv:AgriVLN} benchmark are proposed to pioneeringly extend VLN to the agricultural domain, enabling agricultural robots navigate to the target positions following the natural language instructions. However, when evaluated on the A2A benchmark, AgriVLN only achieves Success Rate (SR) of 0.47 and Navigation Error (NE) of 2.91m, leaving a large gap compared to human performance. 
% Model (VLM)-based
% \cite{TMLR} ,ICRA:Discuss ,ACL:MapGPT
% --- Camera Ready ---
\par In contrast, Vision-and-Language Navigation (VLN) \cite{CVPR:R2R,ECCV:VLN-CE} 
enables navigating robots from starting points to target positions following natural language instructions, 
% enables agents to follow natural language instructions to  
% navigate from starting points to target positions, 
having demonstrated strong performances across various domains, such as R2R \cite{CVPR:R2R} for indoor rooms, TOUCHDOWN \cite{CVPR:TOUCHDOWN} for urban streets, and AerialVLN \cite{ICCV:AerialVLN} for aerial spaces. 
% Motivated by prior Vision-Language Model (VLM)-based methods \cite{AAAI:NavGPT}, 
Inspired by them, 
Zhao et al. \cite{arXiv:AgriVLN} built 
the A2A benchmark 
and 
the AgriVLN method 
to pioneeringly extend VLN to the agricultural domain, thereby effectively alleviating the problems on limited mobility and poor adaptability. 
% enabling agricultural robots navigate to the target positions following the natural language instructions. 
However, when evaluated on A2A, AgriVLN only achieves Success Rate (SR) of 0.47 and Navigation Error (NE) of 2.91 m, leaving a large gap to the performance of human. 
\par We find that one of the biggest challenges stems from the camera height: 
% In the A2A benchmark, all the camera streamings are captured at a height of 0.38m, which better aligns with the practical height of agricultural robot, meanwhile, inevitably constrains the field of visual observation. 
In A2A, all the camera streamings are captured at a height of 0.38 m. On the one hand, it better aligns with the practical height of agricultural robot. On the other hand, it inevitably constrains the field size of visual observation. 
As a classical Chinese poetry suggests, 
\begin{quote}
\textit{“The true face of the mountain remains unseen from within.”}
\end{quote}
agricultural robots tend to attain richer perceptions on immediate surroundings, but less awareness on global spaces.
% contexts
\par To address this limitation, we get inspiration from an interesting daily life phenomenon. Now, imagine that you are travelling in a new city: On the first day, you use a navigation app trying to find your hotel. Although the navigation information is specific enough, it may takes a long time to recognize the landmarks mentioned in the navigation information. On the second day, when you are walking back to the hotel, the navigation information does not change, but it is much easier to find the hotel. We attribute this common 
% example 
phenomenon 
to the subconscious spatial memory: Humans have the ability to spontaneously understand spaces and save them as spatial memories, assisting future movements in same scenes. In contrast, standard VLN methods treat every task as an independent task, overlooking the connection between two tasks in the same scene, as illustrated in Figure \ref{fig:teaser} (a). In the agricultural domain, however, users often repeat one instruction for several times, such as \textit{“Our mate is working in the farm. Go along the path to approach her”}. On a single day, a user may repeat this instruction to ask the agricultural robot to reach the correct position preparing for working for several times. Hence, we highlight a substantial gap between humans and robots on using past spatial memories to assist present episodes.

\par To bridge this gap, we propose the module of \textbf{S}patial \textbf{U}nderstanding \textbf{M}emory (\textbf{SUM}), 
which uses Visual Geometry Grounded Transformer (VGGT) \cite{CVPR:VGGT} to 
execute spatial understanding 
and uses frontal and oblique representations to save spatial memories from the past,  
% in the first turn
thereby assisting the decision-maker to recall the spatial characteristics of the scenes in the present. 
% in the second turn
We integrate our SUM module into the AgriVLN backbone to build the 
method of \textbf{S}patial \textbf{U}nderstanding \textbf{M}emory for \textbf{Agri}cultural \textbf{V}ision-and-\textbf{L}anguage \textbf{N}avigation (\textbf{SUM-AgriVLN}). 
When evaluated on A2A, our SUM-AgriVLN effectively improves SR from 0.47 to 0.54 with only slight sacrifice on NE from 2.91 m to 2.93 m, 
% demonstrating the state-of-the-art performance in the agricultural VLN domain.
surpassing all the previous state-of-the-art agricultural VLN methods \cite{arXiv:AgriVLN,EMNLP:SIA-VLN,RAL:DILLM-VLN}.
Furthermore, we implement the ablation study to discuss the advantages and limitations of SUM, and we implement the case study to show the running process of SUM-AgriVLN.

% \par To bridge this gap, we propose the method of \textbf{S}patial \textbf{U}nderstanding \textbf{M}emory for \textbf{Agri}cultural \textbf{V}ision-and-\textbf{L}anguage \textbf{N}avigation (\textbf{SUM-AgriVLN}). First, we propose the module of \textbf{S}patial \textbf{U}nderstanding \textbf{M}emory (\textbf{SUM}), in which the \textit{Spatial Understanding} step employs Visual Geometry Grounded Transformer (VGGT) \cite{CVPR:VGGT} to reconstruct 3D geometry from sampled camera frames, and the \textit{Spatial Memory} step leverages geometry parsing to extract \textit{frontal} and \textit{oblique} perspective features as spatial memory representations. Second, we integrate the SUM module into the base model of AgriVLN to establish our SUM-AgriVLN method, which loads the 3D reconstruction representation to recall the spatial memory, guiding the understanding on the instruction and camera streaming, to improve the prediction on low-level action sequence. Third, we evaluate SUM-AgriVLN on the A2A benchmark, effectively improving SR from 0.47 to 0.54 with slight sacrifice on NE from 2.91 m to 2.93 m. Compared to the existing VLN methods 
% \cite{arXiv:AgriVLN,EMNLP:SIA-VLN,RAL:DILLM-VLN} 
% evaluated on A2A, our SUM-AgriVLN demonstrates the state-of-the-art performance in the agricultural VLN domain. We further implement the ablation study to show the effectiveness of SUM, and the case study to show the running process of SUM-AgriVLN. 
\par Here we share a simple example to demonstrate the difference between our SUM-AgriVLN and standard agricultural VLN methods, as illustrated in Figure \ref{fig:teaser}. When executing the instruction in the first turn, both methods fail to reach the correct position (SR = 0, NE = 5.27 m). 
When executing the same instruction in the second turn, standard agricultural VLN methods still fail (SR = 0, NE = 5.27 m), while our SUM-AgriVLN 
loads the spatial memory saved from the spatial understanding in the first turn, to assist the decision-maker to recall the scene that this farm \textit{“is a long path leading to our mate”} in the second turn, thereby navigating to a closer position 
% loads the spatial memory from the spatial understanding in the first turn to improve the global recognition, leading to a successful navigation 
(SR = 1, NE = 1.76 m).
\par In summary, our main contributions are as follows:
% Spatial Understanding Memory
\begin{itemize}
\item We propose the SUM module 
% a spatial understanding and spatial memory module, 
to execute spatial understanding via 3D reconstructions and save spatial memories via 2D representations from the past, 
thereby assisting the decision-maker to recall the spatial characteristics of the scenes in the present.
% uses 3D reconstructions to understand spaces then uses 2D representations to memorize them.
% a 3D reconstruction and representation module, which employs spatial understanding and saves spatial memory.
\item We propose the SUM-AgriVLN method integrating the SUM module into the AgriVLN backbone, thereby loading spatial memories to assist navigating agricultural robots from starting points to target positions following natural language instructions.
% \textbf{SUM-AgriVLN}, an agricultural VLN method integrating the SUM module, which loads spatial memories to assist navigating agricultural robots from starting points to target positions following natural language instructions. 
\item We implement the ablation study and the case study discussing the 
% effectiveness 
advantages and limitations 
of our SUM module, and we implement the comparison experiment showing the state-of-the-art agricultural VLN performance of our SUM-AgriVLN method.
\end{itemize}
% \par Codes will be available after acceptance.

\begin{figure*}[t]
\centering
\includegraphics[width=1.0\linewidth]{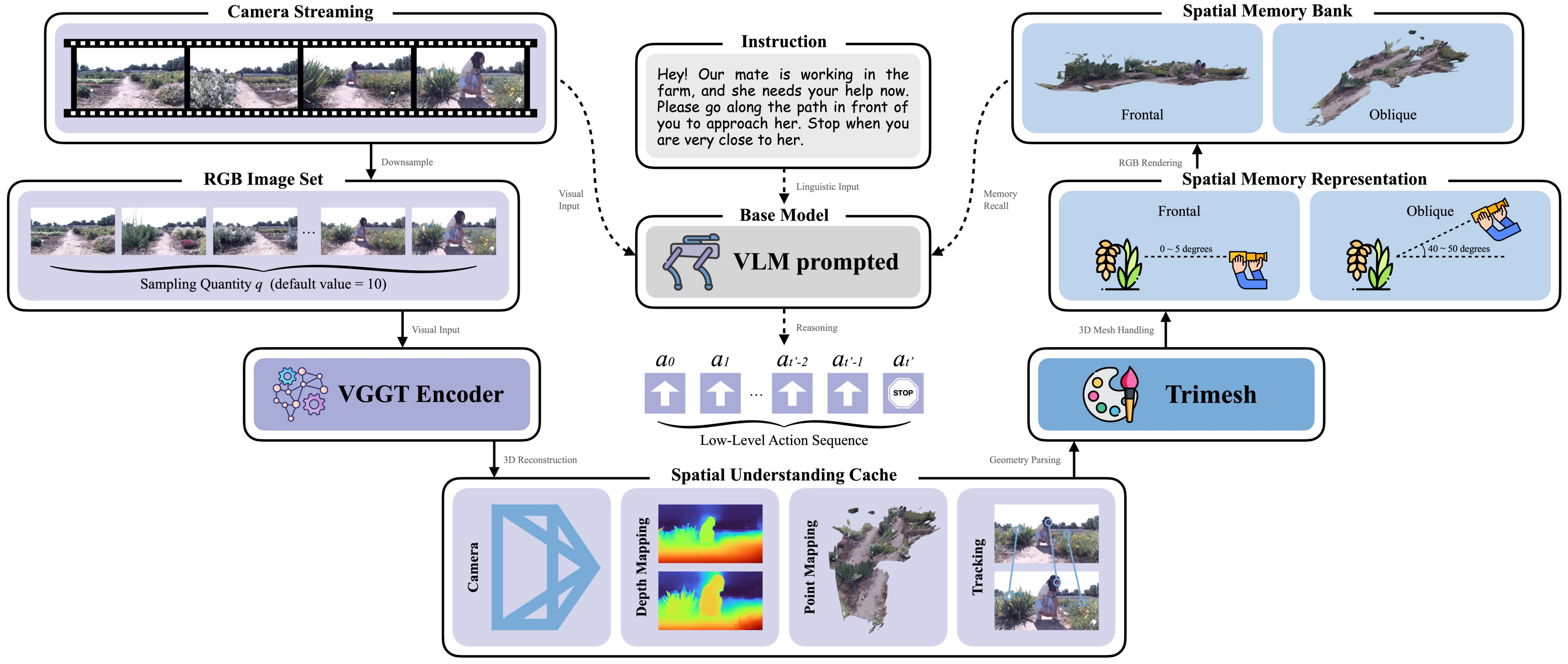}
\caption{SUM-AgriVLN method illustration: 
% The purple and blue parts show the spatial understanding and spatial memory steps, respectively. 
The SUM module executes spatial understanding via 3D reconstructions (marked by purple) and saves spatial memories via 2D representations (marked by blue) from the past, 
% in the first turn
thereby assisting the decision-maker to recall the spatial characteristics of the scenes (marked by grey) in the present.
% in the second turn
% The backbone leverages the spatial memory to recall the scene, thereby better understanding the linguistic and visual inputs to reason the most appropriate low-level action sequence.
}
\label{fig:method}
\end{figure*}

% Section 2
\section{Related Works}
\label{sec:related_works}

\subsection{Agricultural Vision-and-Language Navigation}

\par Agriculture-to-Agriculture (A2A) \cite{arXiv:AgriVLN} is currently the only one VLN benchmark specially designed for agricultural robots, consisting of 1,560 episodes across 6 types of scenes: farm, greenhouse, forest, mountain, garden and village, in which the instructions belong to the step-by-step format and the action space belongs to the continuous environment. 

\par Vision-and-Language Navigation for Agricultural Robots (AgriVLN) \cite{arXiv:AgriVLN} is the first agricultural VLN method, which uses NavGPT \cite{AAAI:NavGPT} as the backbone and integrates the Large Language Model (LLM)-based Subtask List (STL) module, 
enabling navigating agricultural robots from starting points to target positions following natural language instructions.
% enabling an agricultural robot to navigate to a target position following a natural language instruction.

% Traditional VLN methods and benchmarks \cite{CVPR:R2R,ACL:R4R} only focus on indoor scenes, which cannot satisfy the diversified scenarios. 
% - old -
% In recent years, several studies \cite{CVPR:TOUCHDOWN,ICCV:AerialVLN} have been extending VLN to various domains. Nevertheless, none of the benchmarks nor the methods, 
% as far as we know, 
% are designed for agricultural scenes. To bridge this gap, Zhao et al. \cite{arXiv:AgriVLN} propose 
% the A2A benchmark and the AgriVLN method, pioneeringly extending VLN to the agricultural domain. A2A consists of 1,560 episodes across six classifications of scenes: farm, greenhouse, forest, mountain, garden and village, 
% in which all the instructions belong to the step-by-step format and the action space belongs to the continuous environment.
% - old -
% , covering all the common agricultural scenes. 
% When evaluated on A2A, because of the low camera height of 0.38m, AgriVLN robot attains richer perception of the immediate surroundings, but lacks awareness of the global spatial context, leaving a big gap for further improvement.

\subsection{Spatial Memory in Vision-and-Language Navigation}
% ,ICASSP:ERG
In previous VLN methods, 
spatial memories are typically realized through graph-based mapping \cite{ICRA:RoboHop,ICASSP:GAR}, in which environments are abstracted into nodes and edges to facilitate structural reasoning and long-range planning. 
% MapGPT \cite{ACL:MapGPT} leverages map-guided prompting with adaptive path planning to integrate spatial memory into large language models. QueSTMaps \cite{IROS:QueSTMaps} builds queryable semantic topological maps to support structured spatial understanding in 3D environments. MV-Topo \cite{IROS:MV-Topo} enriches topological maps by incorporating multiple visual modalities at each node. 
While these methods advance spatial memory modeling, they remain constrained by explicit graphs or topological structures, which may limits the scalability in large scale environments and the generalization to less structured environments, such as agriculture.

\subsection{3D Reconstruction}
% ,ECCV:SCR CVPR:Cascade, IJCV:Particle,
% ,CVPR:VGGSfM ,CVPR:Rethinking ECCV:PVR,
\par We summarize the existing 3D reconstruction researches into three main directions: Structure from Motion \cite{NeurIPS:DROID-SLAM}, Multi-View Stereo \cite{ECCV:Multiview}, and Point Tracking \cite{CVPR:DOT}. We observe that VGGT \cite{CVPR:VGGT} unifies these advances with a lightweight Transformer, achieving the state-of-the-art reconstruction and tracking while remaining efficient. Hence, we suggest VGGT as an appropriate foundation model for our SUM module.
\section{Methodology}
\label{sec:methodology}
In this section, we present the SUM-AgriVLN method, as illustrated in Figure \ref{fig:method}. First, we introduce the task definition in Sec. \ref{sec:task_definition}. Second, we present the SUM module in Sec. \ref{sec:SUM}. Third, we integrate it into the AgriVLN backbone to build the SUM-AgriVLN method in Sec. \ref{sec:base_model}.
% In this section, we present the method of \textbf{S}patial \textbf{U}nderstanding \textbf{M}emory for \textbf{Agri}cultural \textbf{V}ision-and-\textbf{L}anguage \textbf{N}avigation (\textbf{SUM-AgriVLN}), as illustrated in Figure \ref{fig:method}. First, we present the task definition in Sec. \ref{sec:task_definition}. Second, we present the module of \textbf{S}patial \textbf{U}nderstanding \textbf{M}emory (\textbf{SUM}) in Sec. \ref{sec:SUM}. Third, we integrate the SUM module into the base model to establish the SUM-AgriVLN method in Sec. \ref{sec:base_model}.

\subsection{Task Definition}
\label{sec:task_definition}
The task of Agricultural VLN 
% Vision-and-Language Navigation 
\cite{arXiv:AgriVLN} is defined as follows: In each episode, the model is given an instruction in natural language, denoted as $W = \langle w_1, w_2, \dots, w_L \rangle$, where $L$ is the number of words. At each time step $t$, the model is given a front-facing RGB image $I_t$. The purpose is understanding both $W$ and $I_t$, to select the most appropriate low-level action $\hat{a_t}$ from the action space $\{ \texttt{[FORWARD]}$, $\texttt{[LEFT ROTATE]}$, $\texttt{[RIGHT ROTATE]}$, $\texttt{[STOP]} \}$, thereby navigating the agricultural robot from the starting point to the target position. 

\subsection{Spatial Understanding Memory}
\label{sec:SUM}

We design the SUM module running in a two-step manner: 
% The SUM module consists of two steps: 
the spatial understanding and the spatial memory. 

\subsubsection{Spatial Understanding}

The input is a complete camera image set $\{I_1, I_2, I_3, \dots, I_{t'}\}$ of $I_t \in \mathbb{R}^{3 \times \mathbb{H} \times \mathbb{W}}$. We uniformly sample $q$ frames from the complete camera image set, in which every selected frame is separated by an equal temporal interval $\Delta$. Considering both the accuracy-efficiency balance and the video memory limitation, 
% \footnote{More details are available in Sec. \ref{sec:experimental_settings}.}, 
% Visual Geometry Grounded Transformer 
we set $q$ = 10 as the default value. Next, we introduce VGGT \cite{CVPR:VGGT} as the vision encoder, denoted as $\mathcal{V}\,(\,\cdot\,)$. The sampled camera image set $\{I_1, I_{1+\Delta}, I_{1+2\Delta}, \dots,I_{t'}\}$ is processed to the 3D reconstruction $R$ in the binary glTF format, defined as:
\begin{equation}
R = \mathcal{V} \, (\{I_1, I_{1+\Delta}, I_{1+2\Delta}, \dots, I_{t'}\}, \varphi)
\end{equation}
where $\varphi$ is the set of hyper-parameters for $\mathcal{V}\,(\,\cdot\,)$.
% , detailed in Appendix. 

\subsubsection{Spatial Memory}
We introduce trimesh for geometry parsing and 3D mesh handling, denoted as $\mathcal{T}\,(\,\cdot\,)$, which renders $R$ to point clouds in the 3D RGB representation, denoted as $M$. We manually extract two core perspectives - \textit{frontal} and \textit{oblique} - in the 2D RGB representation, in which the sight angles of \textit{frontal} and \textit{oblique} are set to 0 $\sim$ 5 and 40 $\sim$ 50, respectively. In summary, the whole process of the spatial memory is defined as:
\begin{equation}
M_f, M_o \longleftarrow M = \mathcal{T} \, (R)
\end{equation}
where $M_f, M_o \in \mathbb{R}^{3 \times \mathbb{H}' \times \mathbb{W}'}$ are the spatial memories from the \textit{frontal} and \textit{oblique} perspectives, respectively, in which we set $\mathbb{H}' = 360$ and $\mathbb{W}' = 640$ as the default values. We store $M_f$ and $M_o$ into the spatial memory bank for 
the backbone to load at any time.
% subsequent loading by the base model at any time.

\begin{figure*}[t]
\centering
\includegraphics[width=1.0\linewidth]{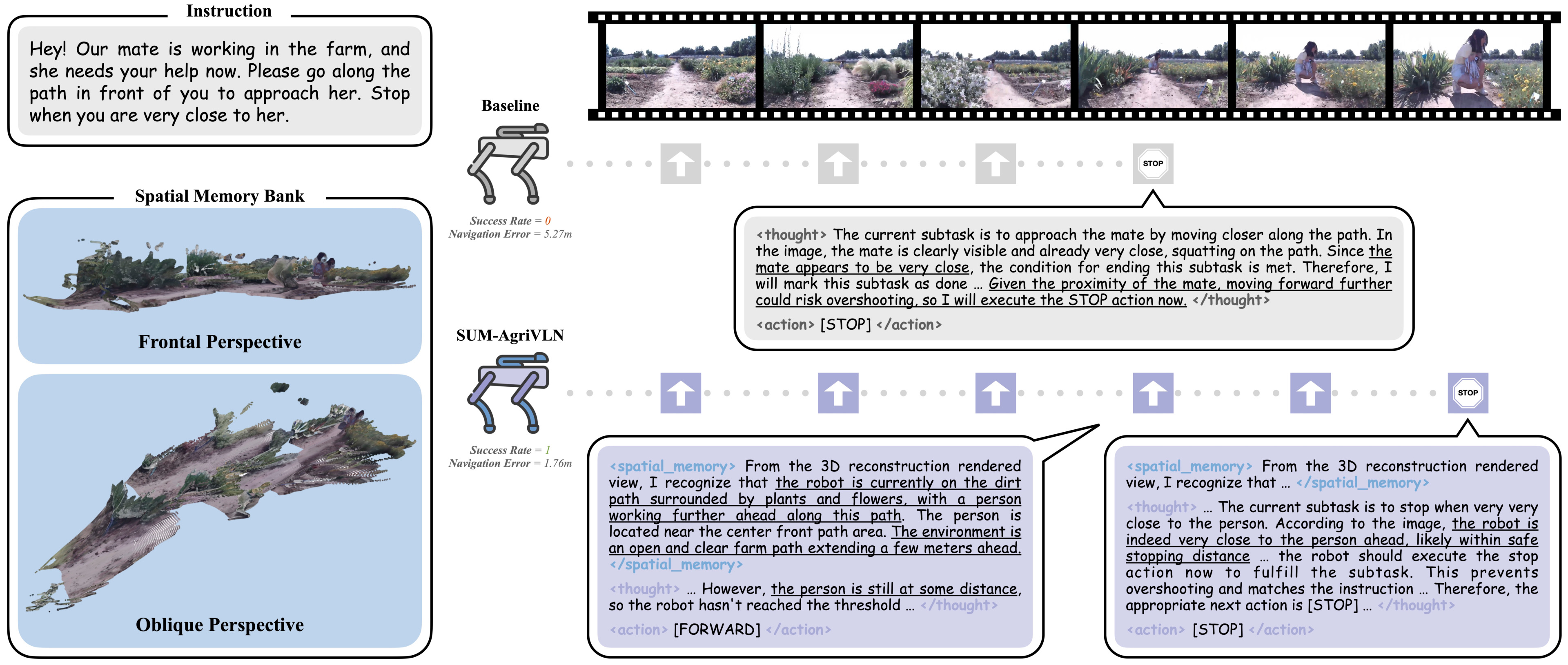}
\caption{Case study illustration: The grey and purple parts show the reasoning thoughts of the baseline and our SUM-AgriVLN, respectively. The blue part shows the spatial memories from the \textit{frontal} and \textit{oblique} perspectives, respectively.}
\label{fig:qualitative_experiment}
\end{figure*}
% (Character's faces are temporarily blurred out to conform to the anonymity principle)
% in 2D representations 

\subsection{Backbone}
\label{sec:base_model}
\par We follow the 
% -based 
Vision-Language Model (VLM)-based
architecture of AgriVLN \cite{arXiv:AgriVLN} as our backbone, denoted as $\mathcal{D}\,(\,\cdot\,)$, in which the SUM module is integrated between the layer of visual input and the module of decision-making. 

% We follow the AgriVLN \cite{arXiv:AgriVLN} method as our backbone. We prompt a Vision-Language Model (VLM) in an end-to-end paradigm, denoted as $\mathcal{D}\,(\,\cdot\,)$.

At each time step $t$, $\mathcal{D}\,(\,\cdot\,)$ loads $\tilde{M} \in \{M_f, M_o, M_f+M_o\}$ to recall the spatial memory to the current scene, then understands both the linguistic input $W$ and the visual input $I_t$, to predict the most appropriate low-level action $\hat{a_t}$ represented inside the linguistic output $l_t$, defined as: 
\begin{equation}
\langle l_1, l_2, \dots, l_{t'} \rangle = \mathcal{D} \, \big(\tilde{M}, \mathcal{S}(W), \langle I_1, I_2, \dots, I_{t'} \rangle, P\big)
\end{equation}
where $\mathcal{S}\,(\,\cdot\,)$ is the STL module. 
% Subtask List
% \footnote{More details are available in AgriVLN \cite{arXiv:AgriVLN}.}. 
$P$ is the prompt template for $\mathcal{D}\,(\,\cdot\,)$. 
% , detailed in Appendix. 
$t'$ denotes the ending time step, at which at least one of the following conditions happens: 
1. $\hat{a_{t'}}$ = $\texttt{[STOP]}$; 
2. The predicted sequence $\langle \hat{a_{t'-\tau}}, \hat{a_{t'-\tau+1}}, \dots, \hat{a_{t'}} \rangle$ is deviated to the label sequence $\langle a_{t'-\tau}, a_{t'-\tau+1}, \dots, a_{t'} \rangle$, where $\tau$ denotes the time step threshold; 
3. $t'$ reaches the upper limitation. In the $\langle l_1, l_2, \dots, l_{t'} \rangle$ sequence, every $l$ consists of several sections wrapped by tag-pair. We use regex match to extract $\hat{a}$ from $l$, defined as:
\begin{equation}
\hat{a}, \rho_m, \rho_d \longleftarrow l
\end{equation}
where $\rho_m$ and $\rho_d$ are the reasoning thoughts on spatial understanding and decision-making, respectively, providing the explicit interpretations.

% Section 4
\section{Experiments}
\label{sec:experiments}
% In this section, we present several experiments from different aspects. First, we present the experimental setting and evaluation metrics in Sec. \ref{sec:experimental_setting} and Sec. \ref{sec:evaluation_metrics}, respectively. Second, we implement the qualitative experiment on the representative episode in Sec. \ref{sec:qualitative_experiment}. Third, we implement the comparison experiment in Sec. \ref{sec:comparison_experiment}. Forth, we implement the ablation experiment in Sec. \ref{sec:ablation_experiment}.  

\subsection{Experimental Settings}
\label{sec:experimental_settings}
We implement all the experiments on the A2A \cite{arXiv:AgriVLN} benchmark, and we access all the LLMs and VLMs through 
% Vision-Language Models
% their corresponding official 
APIs. All the experiments run on a single Apple M1 Pro GPU with 32 G video memory. 
% Specially, in the SUM module, when the VGGT vision encoder is running, the SWAP \cite{DI:SWAP} mechanism of macOS is temporarily activated for an extract video memory.
% (about 5 $\sim$ 20 G).
% \cite{arXiv:Apple_Silicon}

\subsection{Evaluation Metrics}
\label{sec:evaluation_metrics}
We follow the three evaluation metrics in the A2A \cite{arXiv:AgriVLN} benchmark: Success Rate (SR), Navigation Error (NE) and Independent Success Rate (ISR). NE measures the path length between the stopping position and the target position. SR measures the rate successfully reaching the target position within a 2-meter NE. ISR evaluates the success rate of each subtask individually. 
We consider SR as the primary one.
% metric
% We suggest that successfully reaching to the target positions is the priority for navigation, therefore, we select SR as the primary metric.
% \cite{CVPR:ISR}

% Perspective
% experiment 
\begin{table*}[t]
\caption{Comparison results between our SUM-AgriVLN and the state-of-the-art methods on the low-complexity portion (subtask $=$ 2), the high-complexity portion (subtask $\geq$ 3), and the whole of the A2A benchmark. 
\textbf{Bold} and \underline{underline} mark the best and worst scores, respectively.
}
\label{tab:comparison_experiment}
\centering
\renewcommand{\arraystretch}{1.2}
\begin{tabular}{rll ccc|ccc|ccc}
\toprule
\multirow{2}{*}{\textbf{\#}} & \multirow{2}{*}{\textbf{Method}} & \multirow{2}{*}{\textbf{SUM}} & 
\multicolumn{3}{c}{\textbf{A2A ( low-complexity )}} & 
\multicolumn{3}{c}{\textbf{A2A ( high-complexity )}} & 
\multicolumn{3}{c}{\textbf{A2A}} \\
\cmidrule(lr){4-6} \cmidrule(lr){7-9} \cmidrule(lr){10-12}
& & & \textbf{SR} $\uparrow$ & \textbf{NE} $\downarrow$ & \textbf{ISR} & 
\textbf{SR} $\uparrow$ & \textbf{NE} $\downarrow$ & \textbf{ISR} & 
\textbf{SR} $\uparrow$ & \textbf{NE} $\downarrow$ & \textbf{ISR} \\
\midrule
1 & Random   & -                       & 0.13 & 7.30          & - & 0.04 & 6.74 & - & 0.09 & 7.03 & - \\
2 & Fixed    & -                       & 0.00 & 0.00          & - & 0.06 & 6.32 & - & 0.03 & 3.06 & - \\
\midrule \rowcolor{gray!15} \multicolumn{12}{c}{\textit{State-of-the-Art}} \\
3 & SIA-VLN \cite{EMNLP:SIA-VLN}      & - & 0.52             & 1.46             & 3.27 / 3.88 & \underline{0.08} & \underline{5.12} & 2.02 / 4.99 & \underline{0.31} & \underline{3.24} & 2.66 / 4.42 \\
4 & DILLM-VLN \cite{RAL:DILLM-VLN}    & - & \underline{0.41} & \textbf{1.36}    & 4.17 / 5.08 & 0.32             & 3.90             & 2.59 / 4.73 & 0.36             & \textbf{2.60}    & 3.40 / 4.91 \\
5 & AgriVLN \cite{arXiv:AgriVLN}      & - & 0.58             & 2.32             & 2.01 / 2.57 & 0.35             & \textbf{3.54}    & 1.88 / 3.23 & 0.47             & 2.91             & 1.95 / 2.89 \\
\midrule \rowcolor{gray!15} \multicolumn{12}{c}{\textit{Ours}} \\
6 & SUM-AgriVLN$_{f}$ & Frontal           & 0.63             & \underline{2.35} & 2.25 / 2.57 & \textbf{0.43}    & 3.67             & 1.91 / 3.21 & \textbf{0.54}    & 2.99             & 2.09 / 2.88 \\
7 & SUM-AgriVLN$_{o}$ & Oblique           & \textbf{0.66}    & 2.26             & 2.22 / 2.57 & 0.42             & 3.64             & 1.93 / 3.25 & \textbf{0.54}    & 2.93             & 2.08 / 2.90 \\
8 & SUM-AgriVLN$_{h}$ & Frontal + Oblique & 0.65             & 2.14             & 2.23 / 2.60 & 0.40             & 3.66             & 1.94 / 3.24 & 0.53             & 2.87             & 2.09 / 2.91 \\
\midrule
9 & Human    & -                       & 0.93 & 0.32 & - & 0.80 & 0.82 & - & 0.87 & 0.57 & - \\
\bottomrule
\end{tabular}
\end{table*}

\subsection{Case Study}
\label{sec:qualitative_experiment}
% Towards better understanding for readers, 
We implement the case study on a representative episode, as illustrated in Figure \ref{fig:qualitative_experiment}, in which we mark all the core reasoning thoughts with \underline{underline}. 
\par At the time step t = 9.6 s, the baseline thinks that \textit{“the mate appears to be very close”} and \textit{“moving forward further could risk overshooting”}, so chooses to \textit{“execute the STOP action now”}. However, there is still 5.27 m to the target position, resulting in NE = 5.27 m and SR = 0. At the same time step, our SUM-AgriVLN loads the 3D reconstruction representation to recall the spatial memory, thereby generating the macro perception that \textit{“the robot is currently on the dirt path surrounded by plants and flowers, with a person working further ahead along this path”}, which supports making the correct prediction that \textit{“the robot hasn't reached the threshold”}. Until the time step t = 12.4 s, our SUM-AgriVLN thinks that \textit{“the robot is indeed very close to the person ahead, likely within safe stopping distance”}, thereby predicting \texttt{[STOP]} at an appropriate position, leading to NE = 1.76 m and SR = 1.

% \newpage

\subsection{Comparison Experiment}
\label{sec:comparison_experiment}
\par We compare our SUM-AgriVLN with three existing state-of-the-art methods, in which we select the version of \textit{oblique} spatial memory (\#7) as our representative method (will be discussed in Sec. \ref{sec:ablation_experiment_perspective}). Besides, the methods of Random, Fixed and Human are reproduced as the lower and upper bounds, respectively.
% : SIA-VLN \cite{EMNLP:SIA-VLN}, DILLM-VLN \cite{RAL:DILLM-VLN} and AgriVLN \cite{arXiv:AgriVLN}
\par The results of the comparison experiment are shown in Table \ref{tab:comparison_experiment}. On the whole of A2A, our SUM-AgriVLN achieves SR of 0.54 and NE of 2.93 m, surpassing all the previous state-of-the-art methods on SR. On the low-complexity and high-complexity portions, compared to the previous best scores, our SUM-AgriVLN improves SR by 8 and 7 percentage points, respectively, showing the outstanding generalization capability on different complexities. 
\par Meanwhile, we note that our SUM-AgriVLN performs inferiorly on NE. Compared to DILLM-VLN, our SUM-AgriVLN increases NE by 0.90 m and 0.33 m on the low-complexity portion and whole of A2A, respectively. We attribute this sacrifice to the dual character of the SUM module: On the one hand, the spacial memory provides the macroscopic environment cognition, leading to a higher possibility to successfully reach the target with an acceptable error (e.g., SR = 1, NE = 1.80 m). On the other hand, the spatial memory may also limit the potential to recognize the environment by the backbone itself, leading to a lower possibility to successfully reach the target with an excellent error (e.g., SR = 1, NE = 0.60 m) or unsuccessfully reach the target with a collision (i.e., SR = 0, NE = 0.00 m). 
% higher 
% significantly improving SR by 7 percentage points compared to the basline of AgriVLN. 
\par In summary, we suggest that our SUM-AgriVLN achieves the outstanding SR and the acceptable NE, demonstrating the state-of-the-art performance in the agricultural VLN domain.

\subsection{Ablation Study}
\label{sec:ablation_experiment}
% Towards a comprehensive analysis on the proposed SUM module, we implement the ablation experiment from three aspects: rendering perspective, scene classification, and task complexity. 

\subsubsection{Rendering Perspective}
\label{sec:ablation_experiment_perspective}
We ablate the rendering perspective in SUM, as shown in Table \ref{tab:comparison_experiment}, in which we render the spatial memory in the \textit{frontal} (\#6), \textit{oblique} (\#7) and \textit{hybrid} (\#8) perspectives, respectively. In terms of input, \textit{hybrid} combines \textit{frontal} and \textit{oblique}, providing more visual semantic than a single perspective. However, \textit{hybrid} does not show a significant superiority, even performs slightly worse than \textit{frontal} and \textit{oblique} on SR. The possible reason is that single perspective's semantic is abundant enough for the backbone to recall the spatial memory, while multi perspectives' semantic may bring new noises, making the spatial memory become chaotic. Besides, incorporating an additional perspective increases the computational cost, resulting in $\sim$ 18\% growth in the number of extra input tokens. Considering the balance between navigation performance and computational cost, we select SUM-AgriVLN with \textit{oblique} spatial memories as our representative method. 
% In summary,

\begin{table}[t]
\caption{Ablation results on scene classifications. 
% on the A2A benchmark. different 
% experiment 
In every scene classification, \textbf{bold} marks the best score.
}
\label{tab:ablation_experiment_scene}
\centering
\renewcommand{\arraystretch}{1.2}
\begin{tabular}{rl|c|ccc}
\toprule
\# & \textbf{Scene} & \textbf{SUM} & \textbf{SR} $\uparrow$ & \textbf{NE} $\downarrow$ & \textbf{ISR} \\ 
\midrule
10 &            &\ding{55}& 0.66 & 1.79 & 2.13 / 2.73 \\
11 & Farm       & Frontal & \textbf{0.73} & \textbf{1.76} & 2.26 / 2.69 \\
12 &            & Oblique & \textbf{0.73} & 1.80 & 2.26 / 2.76 \\
\midrule
13 &            &\ding{55}& 0.35          & 2.56          & 1.88 / 3.07 \\
14 & Greenhouse & Frontal & \textbf{0.47} & \textbf{2.46} & 1.91 / 3.00 \\
15 &            & Oblique & 0.42          & 2.49          & 1.84 / 3.00 \\
\midrule
16 &            &\ding{55}& 0.33          & 3.68          & 1.78 / 2.86 \\
17 & Forest     & Frontal & 0.34          & 3.97          & 2.02 / 2.88 \\
18 &            & Oblique & \textbf{0.42} & \textbf{3.50} & 2.03 / 2.89 \\
\midrule
19 &            &\ding{55}& 0.50          & 2.84          & 1.94 / 2.88 \\
20 & Mountain   & Frontal & \textbf{0.63} & \textbf{2.28} & 2.34 / 2.84 \\
21 &            & Oblique & 0.59          & 2.94          & 2.19 / 2.81 \\
\midrule
22 &            &\ding{55}& 0.49          & \textbf{3.77} & 2.00 / 3.07 \\
23 & Garden     & Frontal & \textbf{0.56} & 4.26          & 2.05 / 3.09 \\
24 &            & Oblique & \textbf{0.56} & 3.97          & 2.14 / 3.14 \\
\midrule
25 &            &\ding{55}& 0.47          & \textbf{2.92} & 1.93 / 2.73 \\
26 & Village    & Frontal & \textbf{0.53} & 3.24          & 1.73 / 2.73 \\
27 &            & Oblique & \textbf{0.53} & 3.35          & 1.87 / 2.67 \\
\bottomrule
\end{tabular}
\end{table}

\begin{figure*}[t]
\centering
  \begin{subfigure}[b]{0.2\linewidth}
    \includegraphics[width=\linewidth]{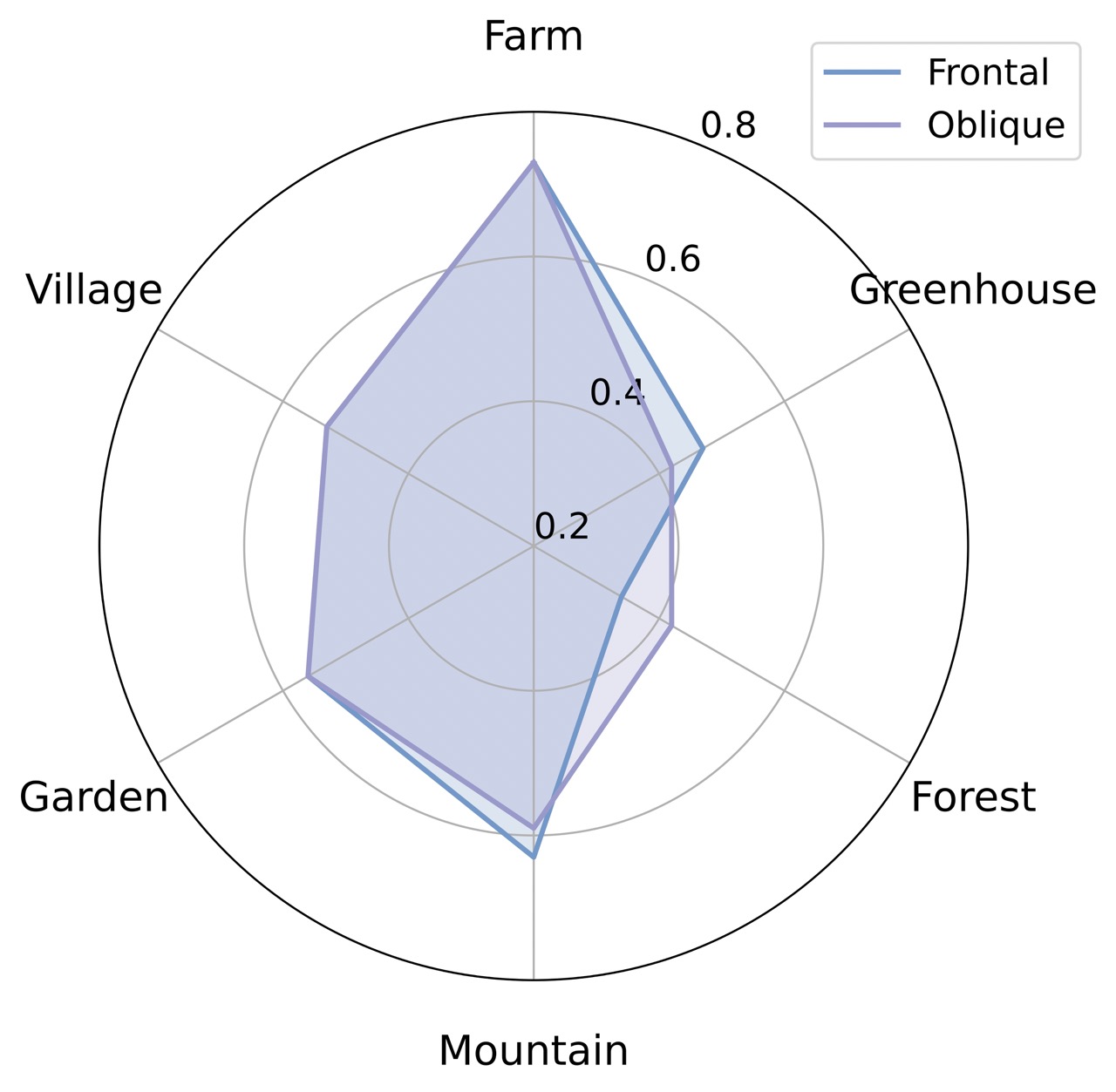}
    \caption{Success Rate $\uparrow$}
  \end{subfigure}
  \hfill
  \begin{subfigure}[b]{0.2\linewidth}
    \includegraphics[width=\linewidth]{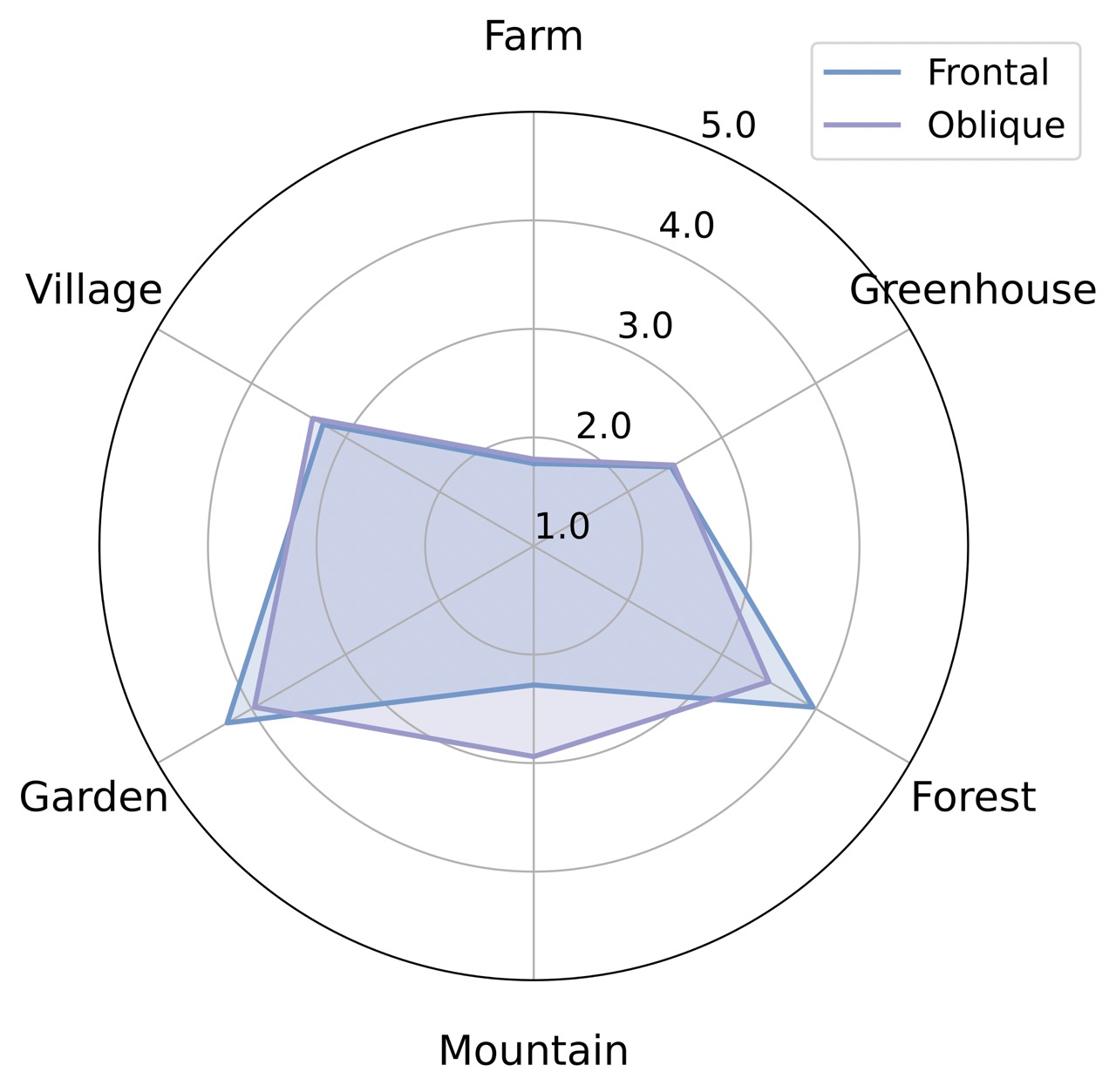}
    \caption{Navigation Error $\downarrow$}
  \end{subfigure}
  \hfill
  \begin{subfigure}[b]{0.27\linewidth}
    \includegraphics[width=\linewidth]{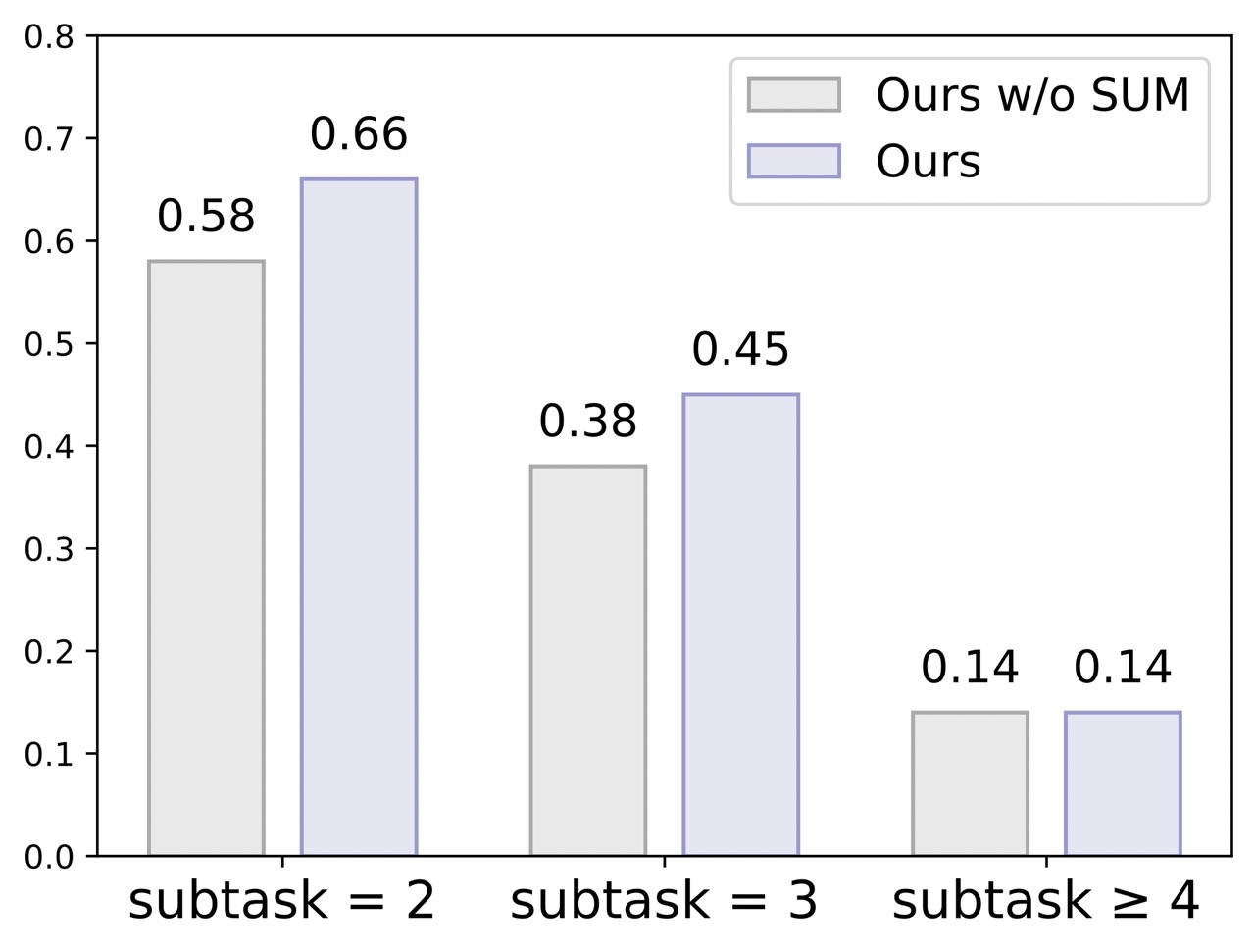}
    \caption{Success Rate $\uparrow$}
  \end{subfigure}
  \hfill
  \begin{subfigure}[b]{0.27\linewidth}
    \includegraphics[width=\linewidth]{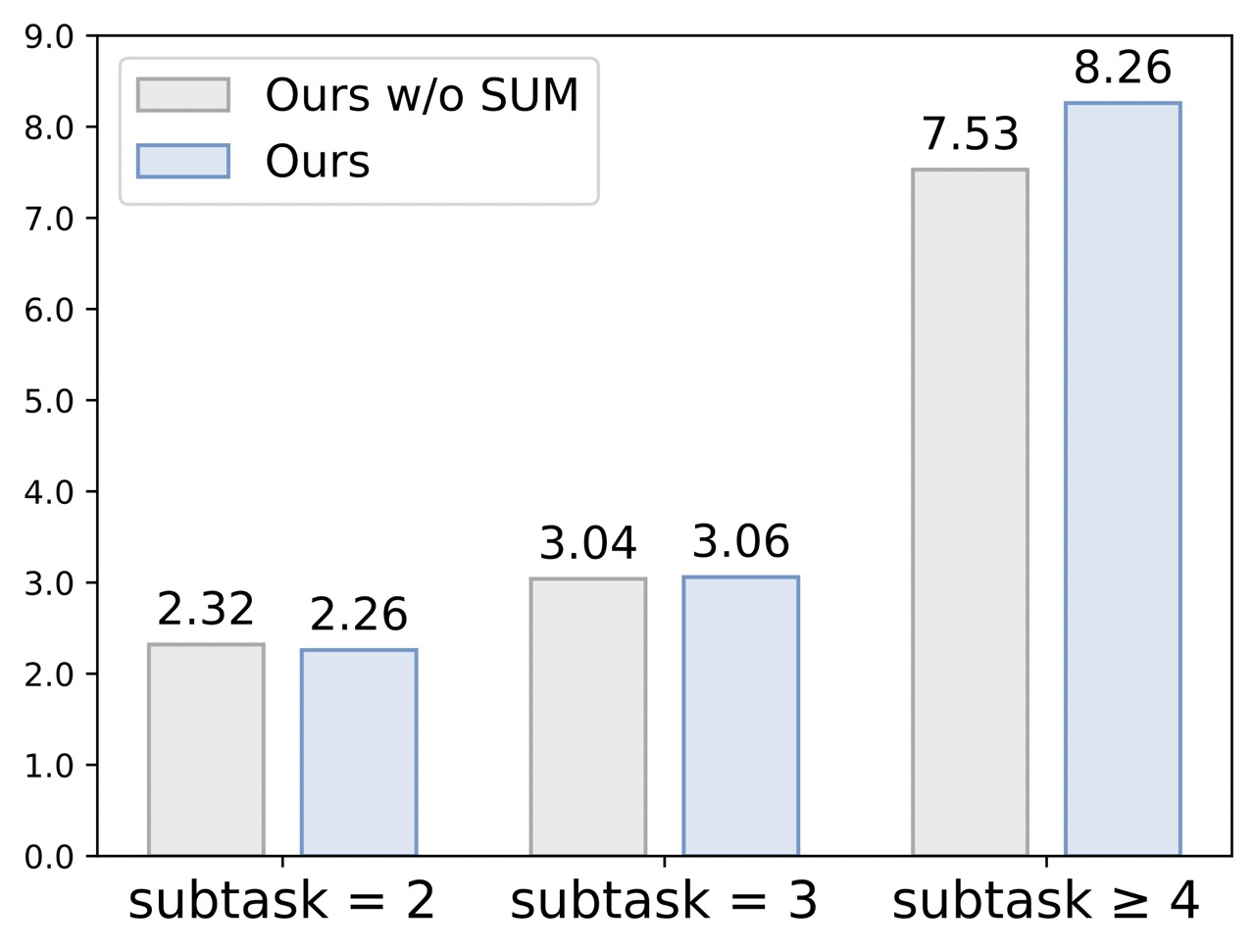}
    \caption{Navigation Error $\downarrow$}
  \end{subfigure}
\caption{Ablation illustrations: (a) and (b) are SUM-AgriVLN with \textit{frontal} v.s. \textit{oblique} spatial memories on different scene classifications. (c) and (d) are SUM-AgriVLN v.s. SUM-AgriVLN $_{w/o}$ SUM on different task complexities.}
\label{fig:ablation_experiment_illustration}
\end{figure*}

\subsubsection{Scene Classification}
\par We ablate SUM from SUM-AgriVLN, as shown in Table \ref{tab:ablation_experiment_scene} and Figure \ref{fig:ablation_experiment_illustration} (a) and (b). 
% (a) and (b). 
% Compared to the baseline model, 
% (\#10, \#13, \#16, \#19, \#22, \#25), the module
When SUM is integrated, the navigation performances improve across all the scene classifications, showing its effectiveness and generalization capability. 
% of the SUM module. 
% \par 
% different
We further analyze the strengths and weaknesses of the two perspectives: \textit{frontal} and \textit{oblique}. In farms (\#11, \#12), gardens (\#23, \#24) and villages (\#26, \#27), they perform similarly. In forests (\#17, \#18) and mountains (\#20, \#21), however, they perform quite differently. Specially, in terms of ISR, they show the similar preferences on completions / decompositions of subtasks (such as 2.02 / 2.88 and 2.03 / 2.89 in forests), from which we suggest the consistent ability of the backbone. Hence, we attribute the gap on SR and NE to the geographic difference: In a forest, the ground is relatively flat, making the object in front block the one behind in the frontal perspective, so \textit{oblique} (\#18) can provide more complete semantic than \textit{frontal} (\#17). In a mountain, however, the ground is relatively fluctuant, making all the objects appear without blocking in the frontal perspective, so \textit{frontal} (\#20) can provide equally complete semantic than \textit{oblique} (\#21). Meanwhile, \textit{frontal} aligns with robot's front-facing camera on the sight angle, 
making the spatial memory easier to recall.
% making the spatial memory to be recalled more easily and precisely.

\subsubsection{Task Complexity}
\par We ablate SUM to further analyze the effectiveness on different task complexities, as shown in Table \ref{tab:ablation_experiment_subtask_quantity} and Figure \ref{fig:ablation_experiment_illustration} (c) and (d), in which we follow SUM-AgriVLN with \textit{oblique} spatial memories as our representative method. On the low-complexity portion (subtask = 2) of A2A, when SUM is integrated (\#28, \#31), SR increases from 0.58 to 0.66 and NE decreases from 2.32 m to 2.26 m, bringing a substantial improvement. However, as the task complexity increases, the improvement becomes less and less significant. When the subtask quantity is more than 4 (\#30, \#33), the performance becomes even worse, in which SR keeps 0.14 while NE increases from 7.53 m to 8.26 m. Hence, we suggest that SUM plays a more important role in low-complexity tasks than in high-complexity tasks. 
% \par We attribute this ...
% $\geq$  benchmark

% Section 5
\section{Conclusion}
\label{sec:conclusion}

In this paper, 
we propose the SUM module, which executes spatial understanding via 3D reconstructions and saves spatial memories via 2D representations from the past, thereby assisting the decision-maker to recall the spatial characteristics in the present. We integrate it into the AgriVLN backbone to build the SUM-AgriVLN method, 
% of the scenes
% . When evaluated on A2A, it effectively improves SR from 0.47 to 0.54 with slight sacrifice on NE from 2.91 m to 2.93 m, showing 
achieving the state-of-the-art performance in the agricultural VLN domain.

% In this paper, we propose the SUM-AgriVLN method, in which the SUM module employs spatial understanding and save spatial memory through 3D reconstruction and representation. When evaluated on A2A, our SUM-AgriVLN effectively improves SR from 0.47 to 0.54 with slight sacrifice on NE from 2.91 m to 2.93 m, demonstrating the state-of-the-art performance in the agricultural VLN domain.

\par During the experiments, we find two limitations: 1) The model is currently restricted to handling static scenes. 
% When dynamic objects appear in the video streaming, the 3D reconstruction tends to produce ghosting effects, which degrade the quality of spatial memory. 
2) The spatial memory is represented using 2D RGB images, which inherently offer limited capacity for encoding spatial information, leading to the loss of key features generated during the spatial understanding process. 
% 3) The model requires an initial pre-exploration phase to gather necessary spatial understanding.
\par In the future, in addition to the improvements on the existing limitations, we plan to explore would agricultural VLN methods perform better if we could give them extra depth features, and design a new method to represent depth features from monocular RGB images.
% we plan further researching the dynamic spatial understanding from real-time camera streaming, leading to the better generalization ability.

\section*{Acknowledgment}
This work is supported by the Sichuan Chengdu Modern Agricultural Industry Research Institute of China Agricultural University: Provincial and Municipal Agricultural Subsidy Funded Project; the Natural Science Foundation of Sichuan Province (2024NSFSC0389); and the Provincial and Municipal Agricultural Subsidy Special Funds for the Construction of CAU–SCCD Advanced Agricultural and Industrial Institute. 
This work is also supported by Xiaobei Zhao. 
Thanks to Bangkok, Datong, Beijing, Samarkand, Kathmandu and Pokhara for the impressive traveling experiences, giving us a chilled vibe for experiment and writing. Thanks to Yuanquan Xu, the inspiration to us.

\begin{table}[t]
\caption{Ablation results on task complexities. 
% on the A2A benchmark. experiment different 
In every subtask quantity, 
\textbf{bold} and \underline{underline} mark better and worse scores when SUM is integrated, respectively.
% compared to ours $_{w/o}$ SUM, respectively.
% \textbf{Bold} and \underline{underline} mark better and worse scores compared to the baseline model, respectively.
}
\label{tab:ablation_experiment_subtask_quantity}
\centering
\renewcommand{\arraystretch}{1.2}
\begin{tabular}{rl|c|ccc}
\toprule
\# & \textbf{Method} & \textbf{Subtask} & \textbf{SR} $\uparrow$ & \textbf{NE} $\downarrow$ & \textbf{ISR} \\ 
\midrule
28 &                   & 2       & \textbf{0.66} & \textbf{2.26}    & 2.22 / 2.57 \\
29 & Ours              & 3       & \textbf{0.45} & \underline{3.06} & 1.95 / 3.10 \\
30 &                   & $\geq$4 & 0.14          & \underline{8.26} & 1.79 / 4.43 \\
\midrule
31 &                   & 2       & 0.58 & 2.32 & 2.01 / 2.58 \\
32 & Ours $_{w/o}$ SUM & 3       & 0.38 & 3.04 & 1.92 / 3.10 \\
33 &                   & $\geq$4 & 0.14 & 7.53 & 1.57 / 4.29 \\
\bottomrule
\end{tabular}
% \vspace{-0.35cm}
\end{table}

% references

\end{document}